\documentclass[11pt,a4paper]{article}
\usepackage[hyperref]{acl2019}
\usepackage{times}
\usepackage{latexsym}

\usepackage{graphicx} 
\usepackage{url}
\usepackage{bm}
\usepackage{amsfonts}
\usepackage{subcaption}
\usepackage{amsmath}
\usepackage{comment}
\usepackage{multirow}
\usepackage{array}
\usepackage{float}
\usepackage{CJKutf8}
\usepackage{stfloats}

\makeatletter
\newcommand{\thickhline}{%
\noalign {\ifnum 0=`}\fi \hrule height 0.8pt
\futurelet \reserved@a \@xhline
}
\newcolumntype{"}{@{\hskip\tabcolsep\vrule width 1pt\hskip\tabcolsep}}
\makeatother

\aclfinalcopy 

\title{Multiple Character Embeddings for Chinese Word Segmentation}

\author{
Jingkang Wang\thanks{\ \ \ Equal contribution   (alphabetical order).} \quad  Jianing Zhou$^{*}$ \quad Jie Zhou \quad Gongshen Liu\thanks{\ \ \ Corresponding author.} \vspace{1.5mm} \\
      The Lab of Information Content Intelligent Analysis, Shanghai, China \\
      School of Cyber Science and Engineering, Shanghai Jiao Tong University\\
      \texttt{\{wangjksjtu,zhjjn1919\}@gmail.com, \{sanny02,lgshen\}@sjtu.edu.cn}
}

\begin{document}
\begin{CJK*}{UTF8}{gbsn}
\maketitle
\begin{abstract}
    Chinese word segmentation (CWS) is often regarded as a character-based sequence labeling task in most current works which have achieved great success with the help of powerful neural networks. However, these works neglect an important clue: \textit{Chinese characters incorporate both semantic and phonetic meanings}. In this paper, we introduce multiple character embeddings including \textit{Pinyin Romanization} and \textit{Wubi Input}, both of which are easily accessible and effective in depicting semantics of characters. We propose a novel \textit{shared Bi-LSTM-CRF} model to fuse linguistic features efficiently by sharing the LSTM network during the training procedure. Extensive experiments on five corpora show that extra embeddings help obtain a significant improvement in labeling accuracy. Specifically, we achieve the state-of-the-art performance in AS and CityU corpora with F1 scores of 96.9 and 97.3, respectively without leveraging any external lexical resources.
\end{abstract}

\section{Introduction}
Chinese is written without explicit word delimiters so word segmentation (CWS) is a preliminary and essential pre-processing step for most natural language processing (NLP) tasks in Chinese, such as part-of-speech tagging (POS) and named-entity recognition (NER). The representative approaches are treating CWS as a character-based sequence labeling task following \citet{xue2003chinese} and \citet{Peng2004}. 

Although not relying on hand-crafted features, most of the neural network models rely heavily on the embeddings of characters. Since \citet{DBLP:journals/corr/MikolovSCCD13} proposed word2vec technique, the vector representation of words or characters has become a prerequisite for neural networks to solve NLP tasks in different languages.


However, existing approaches neglect an important fact that Chinese characters contain both semantic and phonetic meanings - there are various representations of characters designed for capturing these features. The most intuitive one is \textit{Pinyin Romanization} (拼音) that keeps many-to-one relationship with Chinese characters - for one character, different meanings in specific context may lead to different pronunciations. This phenomenon called \textit{Polyphony} (and \textit{Polysemy}) in linguistics is very common and crucial to word segmentation task. Apart from Pinyin Romanization, \textit{Wubi Input} (五笔) is another effective representation which absorbs semantic meanings of Chinese characters. Compared to Radical (偏旁) \cite{DBLP:journals/corr/SunLTYJW14,101007,DBLP:conf/ijcnlp/ShaoHTN17}, Wubi includes more comprehensive graphical and structural information that is highly relevant to the semantic meanings and word boundaries, due to plentiful pictographic characters in Chinese and effectiveness of Wubi in embedding the structures.

This paper will thoroughly study how important the extra embeddings are and what scholars can achieve by combining extra embeddings with representative models. To leverage extra phonetic and semantic information efficiently, we propose a shared Bi-LSTMs-CRF model, which feeds embeddings into three stacked LSTM layers with shared parameters and finally scores with CRF layer. We evaluate the proposed approach on 
five corpora and demonstrate that our method produces state-of-the-art results and is highly efficient as 
previous single-embedding scheme. 


Our contributions are summarized as follows: 1) We firstly propose to leverage both semantic and phonetic features of Chinese characters in NLP tasks by introducing Pinyin Romanization and Wubi Input embeddings, which are easily accessible and effective in representing semantic and phonetic features; 2) We put forward a \textit{shared Bi-LSTM-CRF} model for efficiently integrating multiple embeddings and sharing useful linguistic features; 3) We evaluate the proposed multi-embedding scheme on Bakeoff2005 and CTB6 corpora. Extensive experiments show that auxiliary embeddings help achieve state-of-the-art performance without external lexical resources.

\begin{figure}[t]
\centering
	\includegraphics[width=0.49\textwidth]{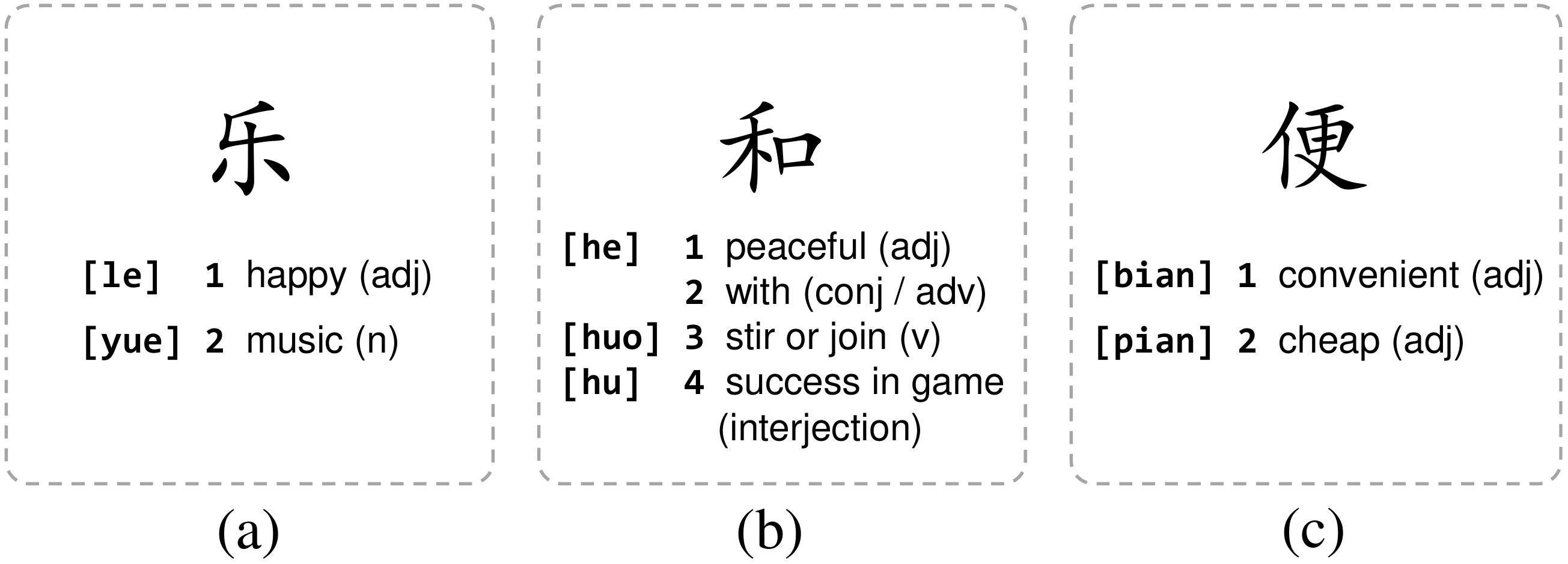}
	\caption{Examples of phono-semantic compound characters and polyphone characters.}
	\label{fig:pinyin}
\end{figure}

\begin{figure}[t]
	\centering
	\includegraphics[width=0.48\textwidth]{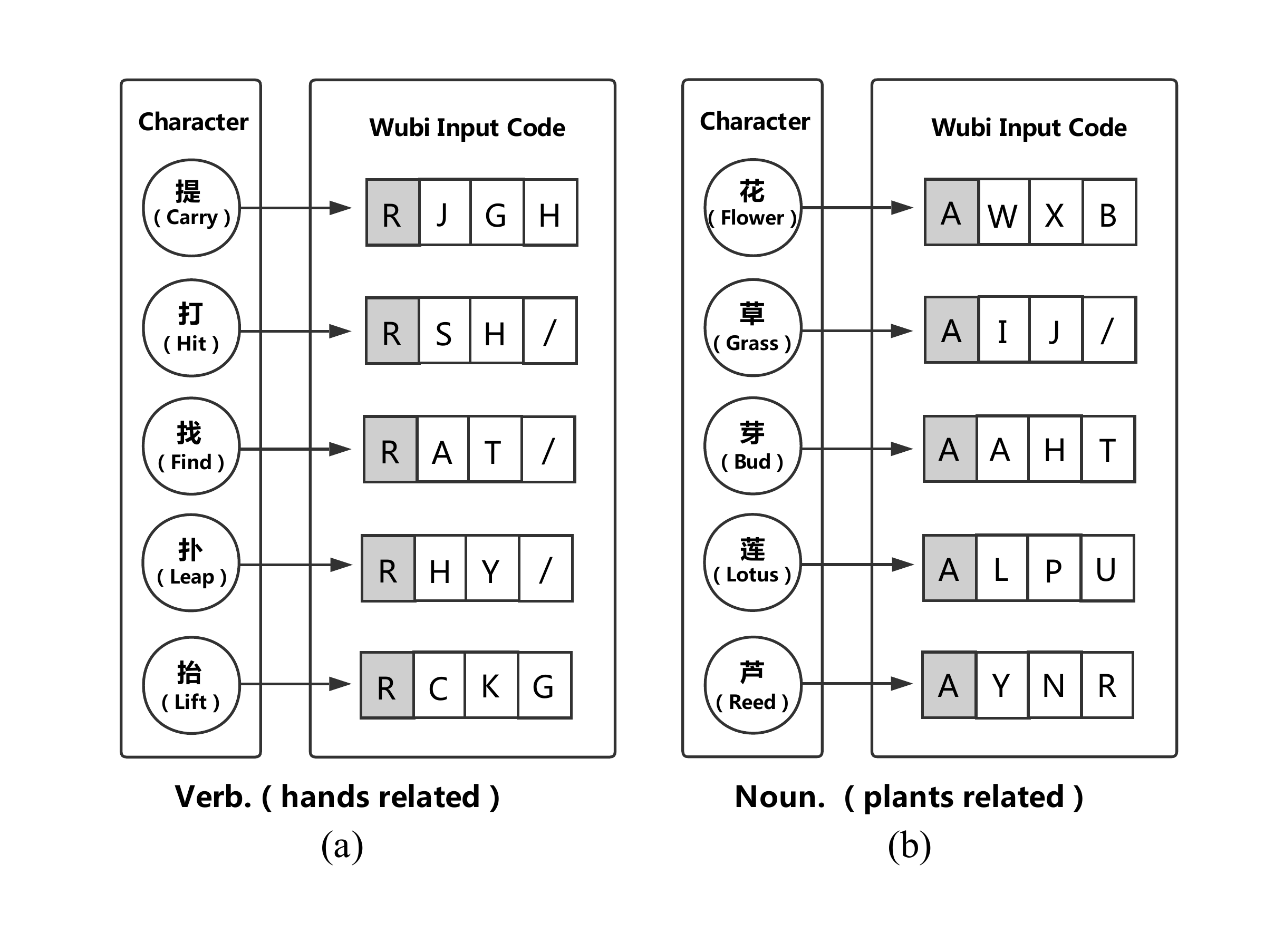}
	\vspace{-5mm}
	\caption{Potential semantic relationships between Chinese characters and Wubi Input. Gray area indicates that these characters have the same first letter in the Wubi Input representation.}
	\label{fig:wubi}
	\vspace{-5mm}
\end{figure}

\section{Multiple Embeddings}
To fully leverage various properties of Chinese characters, we propose to split the character-level embeddings into three parts: character embeddings for textual features, Pinyin Romanization embeddings for phonetic features and Wubi Input embeddings for structure-level features. 

\subsection{Chinese Characters}


CWS is often regarded as a character-based sequence labeling task, which aims to label every character with \{\textit{B, M, E, S}\} tagging scheme. Recent studies show that character embeddings are the most fundamental inputs for neural networks~\cite{DBLP:conf/emnlp/ChenQZLH15,DBLP:conf/acl/CaiZ16,DBLP:conf/acl/CaiZZXWH17}. However, Chinese characters are developed to absorb and fuse phonetics, semantics, and hieroglyphology. In this paper, we would like to explore other linguistic features so the characters are the basic inputs with two other presentations (\textit{Pinyin} and \textit{Wubi}) introduced as auxiliary.



\subsection{Pinyin Romanization}
\textit{Pinyin Romanization} (拼音) is the official romanization system for standard Chinese characters~\cite{ISO7098}, representing the pronunciation of Chinese characters like phonogram in English. Moreover, Pinyin is highly relevant to semantics - one character may correspond varied Pinyin code that indicates different semantic meanings. This phenomenon is very common in Asian languages and termed as polyphone. 

Figure~\ref{fig:pinyin} shows several examples of polyphone characters. For instance, the character `乐' in Figure~\ref{fig:pinyin} (a) has two different pronunciations (Pinyin code). When pronounced as `yue', it means 'music', as a noun. However, with the pronunciation of 'le',  it refers to 'happiness'. Similarly, the character `和' in Figure~\ref{fig:pinyin} (b) even has four meanings with three varied Pinyin code.

Through Pinyin code, a natural bridge is constructed between the words and their semantics. Now that human could understand the different meanings of characters according to varied pronunciations, the neural networks are also likely to learn the mappings between semantic meanings and Pinyin code automatically.

Obviously, Pinyin provides extra phonetic and semantic information required by some basic tasks such as CWS. It is worthy to notice that Pinyin is a dominant computer input method of Chinese characters, and it is easy to represent characters with Pinyin code as supplementary inputs.



\subsection{Wubi Input}
\label{sec:wubi_input}
\textit{Wubi Input} (五笔) is based on the structure of characters rather than the pronunciation. Since plentiful Chinese characters are hieroglyphic, Wubi Input can be used to find out the potential semantic relationships as well as the word boundaries. It is beneficial to CWS task mainly in two aspects: 1) Wubi encodes high-level semantic meanings of characters; 2) characters with similar structures (e.g., radicals) are more likely to make up a word, which effects the word boundaries.

To understand its effectiveness in structure description, one has to go through the rules of Wubi Input method. It is an efficient encoding system which represents each Chinese character with at most four English letters. Specifically, these letters are divided into five regions, each of which represents a type of structure (stroke, 笔画) in Chinese characters. 

Figure~\ref{fig:wubi} provides some examples of Chinese characters and their corresponding Wubi code (four letters).
For instance, `提' (carry), `打' (hit) and `抬' (lift) in Figure~\ref{fig:wubi} (a) are all verbs related to hands and correspond different spellings in English. On the contrary, in Chinese, these characters are all left-right symbols and have the same radical (`R' in Wubi code). That is to say, Chinese characters that are highly semantically relevant usually have similar structures which could be perfectly captured by Wubi. Besides, characters with similar structures are more likely to make up a word. For example, `花' (flower), `草' (grass) and `芽' (bud) in Figure~\ref{fig:wubi} (b) are nouns and represent different plants. Whereas, they are all up-down symbols and have the same radical (`A' in Wubi code). These words usually make up new words such as `花草' (flowers and grasses) and `花芽' (the buds of flowers).

\begin{figure*}[t!]
\centering
\begin{subfigure}[b]{0.27\linewidth}
\includegraphics[width=\textwidth]{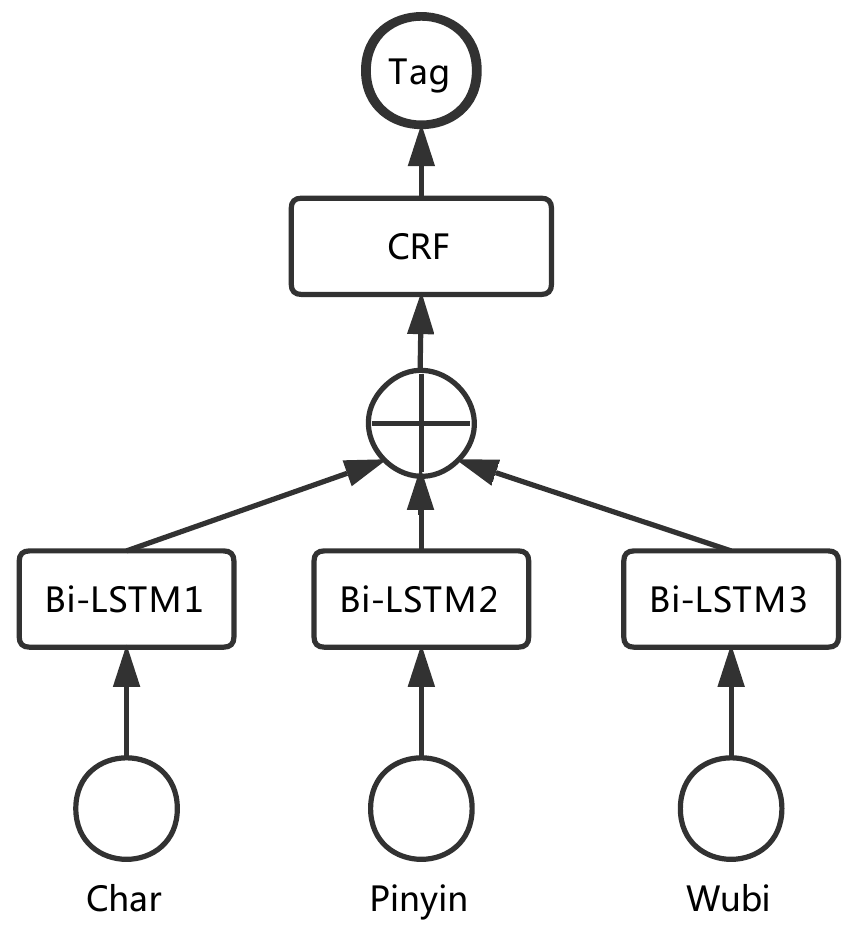}
\caption{Model-I}\label{fig:model1}
\end{subfigure}
\hspace{4mm}
\begin{subfigure}[b]{0.24\linewidth}
\includegraphics[width=\textwidth]{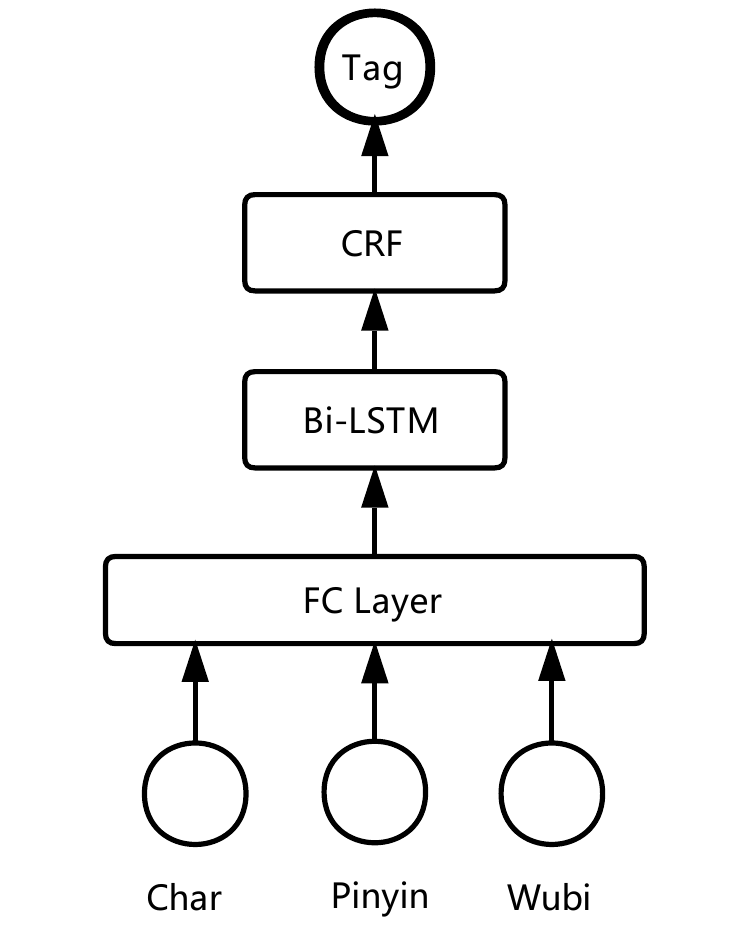}
\caption{Model-II}\label{fig:model2}
\end{subfigure}
\hspace{4mm}
\begin{subfigure}[b]{0.25\linewidth}
\includegraphics[width=\textwidth]{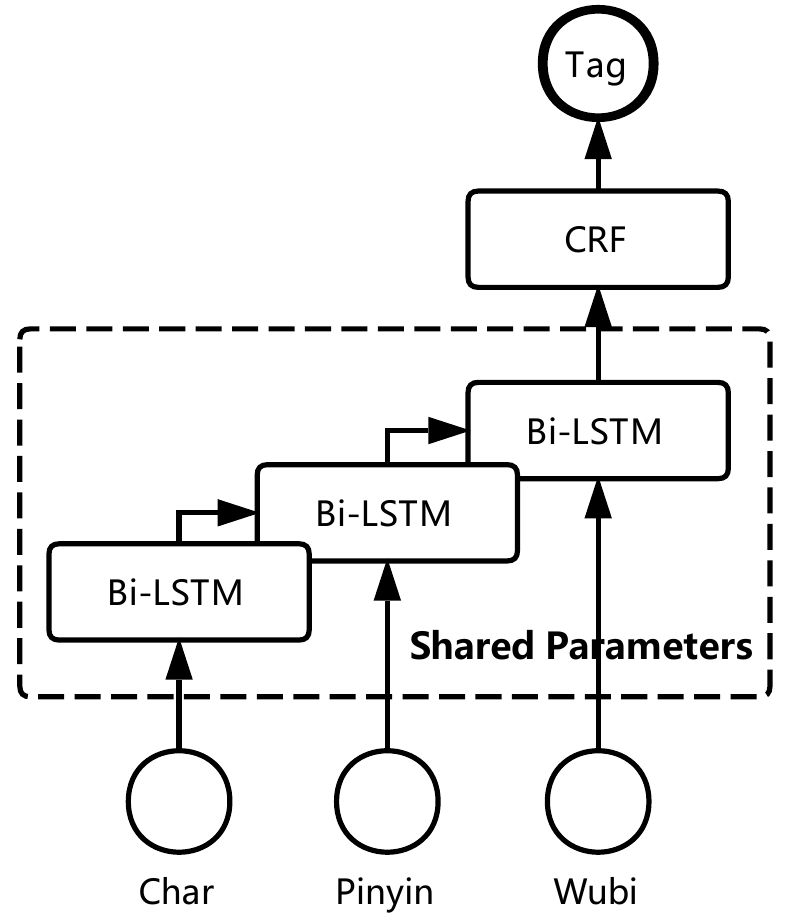}
\caption{Model-III}\label{fig:model3}
\end{subfigure}
\caption{Network architecture of three multi-embedding models.  (a) Model-I: Multi-Bi-LSTMs-CRF Model. (b) Model-II: FC-Layer Bi-LSTMs-CRF Model. (c) Model-III: Shared Bi-LSTMs-CRF Model.}
\label{fig:architecture}
\vspace{-2mm}
\end{figure*}


In addition, the sequence in Wubi code is one approach to interpret the relationships between Chinese characters. In Figure~\ref{fig:wubi}, it is easy to find some interesting component rules. For instance, we can conclude: 1) the sequence order implies the order of character components (e.g., `IA' vs `AI' and `IY' vs `YI'); 2) some code has practical meanings (e.g., `I' denotes water). Consequently, Wubi is an efficient encoding of Chinese characters so incorporated 
as a supplementary input like Pinyin in our multi-embedding model.

\subsection{Multiple Embeddings}
To fully utilize various properties of Chinese characters, we construct the Pinyin and Wubi embeddings as two supplementary character-level features. We firstly pre-process the characters and obtain the basic character embedding following the strategy in \citet{lample2016neural,DBLP:conf/ijcnlp/ShaoHTN17}. Then we use the Pypinyin Library\footnote{\url{https://pypi.python.org/pypi/pypinyin}} to annotate Pinyin code, and an official transformation table\footnote{\url{http://wubi.free.fr/index\_en.html}} to translate characters to Wubi code. Finally, we retrieve multiple embeddings using word2vec tool~\cite{DBLP:journals/corr/MikolovSCCD13}. 

For simplicity, we treat Pinyin and Wubi code as units like characters processed by canonical word2vec, which may discard some semantic affinities. It is worth noticing that the sequence order in Wubi code is an intriguing property considering the fact that structures of characters are encoded by the order of letters (see Sec~\ref{sec:wubi_input}). This point merits further study. Finally, we remark that generating Pinyin code relies on the external resources (statistics prior). Nontheless, Wubi code is converted under a transformation table so does not introduce any external resources. 
\section{Multi-Embedding Model Architecture} \label{sec:architecture}
\begin{figure}[t!]
	\centering
	\includegraphics[width=0.48\textwidth]{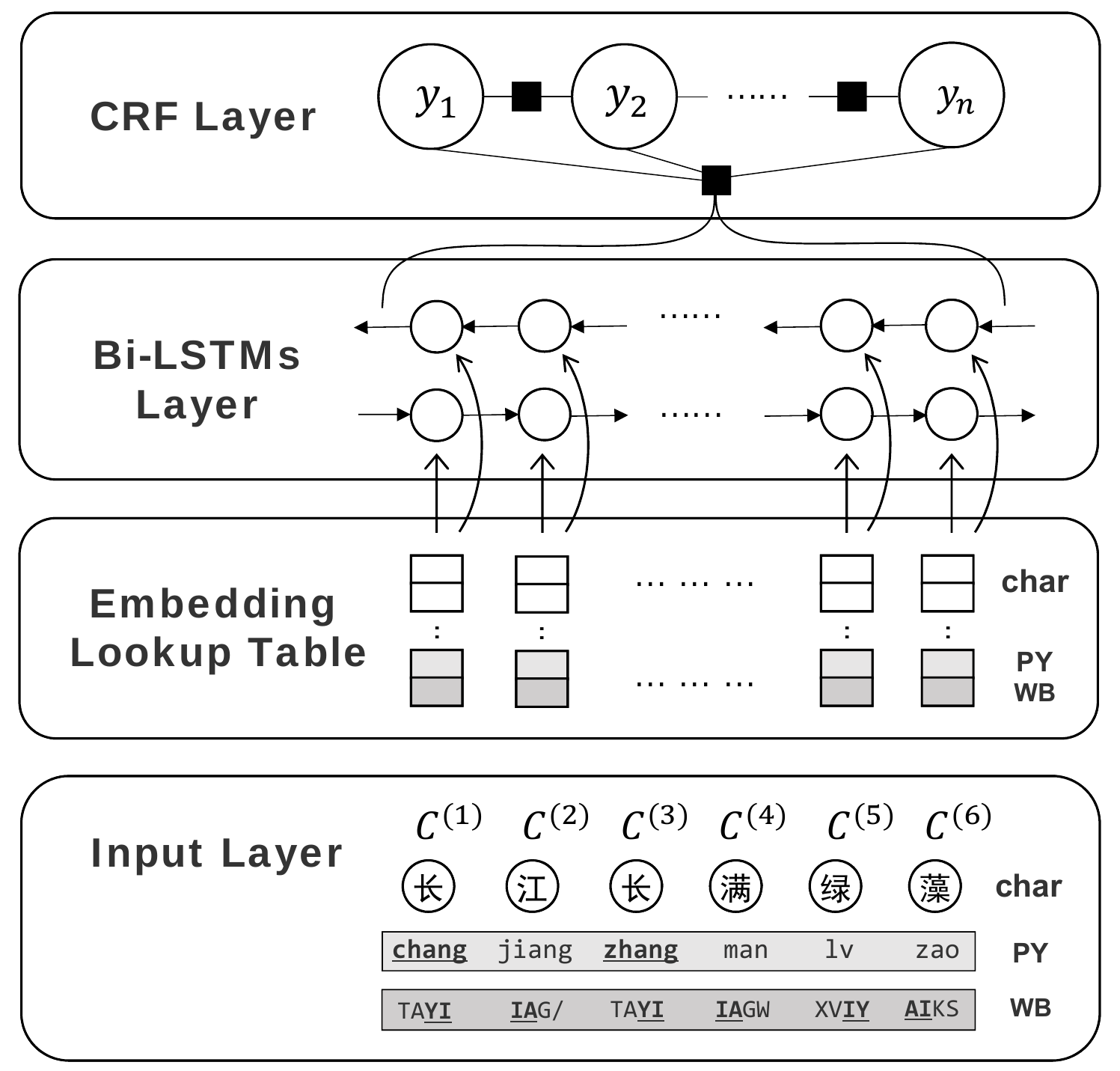}
	\vspace{-0.5cm}
	\caption{The architecture of Bi-LSTM-CRF network. PY and WB represent \textit{Pinyin Romanization} and \textit{Wubi Input} introduced in this paper.}
	\label{fig:baseline}
	\vspace{-0.3cm}
\end{figure}

We adopt the popular Bi-LSTMs-CRF as our baseline model (Figure~\ref{fig:baseline} without Pinyin and Wubi input), similar to the architectures proposed by \citet{lample2016neural} and \citet{101007}. To obtain an efficient fusion and sharing mechanism for multiple features, we design three varied architectures (see Figure~\ref{fig:architecture}). In what follows, we will provide detailed explanations and analysis.

\subsection{Model-\uppercase\expandafter{\romannumeral1}: Multi-Bi-LSTMs-CRF Model}
In Model-\uppercase\expandafter{\romannumeral1} (Figure~\ref{fig:model1}), the input vectors of character, pinyin and wubi embeddings are fed into three independent stacked Bi-LSTMs networks and the output high-level features are fused via addition:
\begin{equation}
\begin{aligned}
\mathbf{h}_{3,c}^{(t)} &= \textrm{Bi-LSTMs}_{1}(\mathbf{x}_{c}^{(t)}, \theta_{c}), \\
\mathbf{h}_{3,p}^{(t)} &= \textrm{Bi-LSTMs}_{2}(\mathbf{x}_{p}^{(t)}, \theta_{p}), \\
\mathbf{h}_{3,w}^{(t)} &= \textrm{Bi-LSTMs}_{3}(\mathbf{x}_{w}^{(t)}, \theta_{w}), \\
\mathbf{h}^{(t)} &= \mathbf{h}_{3,c}^{(t)} + \mathbf{h}_{3,p}^{(t)} + \mathbf{h}_{3,w}^{(t)}, 
\end{aligned}
\label{eq:model2}
\end{equation}
where $\theta_{c}$, $\theta_{p}$ and $\theta_{w}$ denote parameters in three Bi-LSTMs networks respectively. The outputs of three-layer Bi-LSTMs are $\mathbf{h}_{3,c}^{(t)}$, $\mathbf{h}_{3,p}^{(t)}$ and $\mathbf{h}_{3,w}^{(t)}$, which form the input of the CRF layer $\mathbf{h}_{(t)}$. Here three LSTM networks maintain independent parameters for multiple features thus leading to a large computation cost during training.


\subsection{Model-\uppercase\expandafter{\romannumeral2}: FC-Layer Bi-LSTMs-CRF Model}
On the contrary, Model-\uppercase\expandafter{\romannumeral2} (Figure~\ref{fig:model2}) incorporates multiple raw features directly by inserting one fully-connected (FC) layer 
to learn a mapping between fused linguistic features and concatenated raw input embeddings.
Then the output of this FC layer is fed into the LSTM network:
\begin{equation}
\begin{aligned}
\mathbf{x}_{in}^{(t)} &= [\mathbf{x}_{c}^{(t)};\mathbf{x}_{p}^{(t)};\mathbf{x}_{w}^{(t)}], \\
\mathbf{x}^{(t)} &= \sigma(\mathbf{W}_{fc}\mathbf{x}_{in}^{(t)} + \mathbf{b}_{fc}),
\end{aligned}
\end{equation}
where $\sigma$ is the logistic sigmoid function; $\mathbf{W}_{fc}$ and $\mathbf{b}_{fc}$ are trainable parameters of fully connected layer; $\mathbf{x}_{c}^{(t)}$, $\mathbf{x}_{p}^{(t)}$ and $\mathbf{x}_{w}^{(t)}$ are the input vectors of character, pinyin and wubi embeddings.
The output of the fully connected layer $\mathbf{x}^{(t)}$ forms the input sequence of the Bi-LSTMs-CRF. 
This architecture benefits from its low computation cost but suffers from insufficient extraction from raw code. Meanwhile, Model-\uppercase\expandafter{\romannumeral1} and Model-\uppercase\expandafter{\romannumeral2} ignore the interactions between different embeddings. 

\subsection{Model-\uppercase\expandafter{\romannumeral3}: Shared Bi-LSTMs-CRF Model}
To address feature dependency while maintaining training efficiency, Model-\uppercase\expandafter{\romannumeral3} (Figure~\ref{fig:model3}) introduces a sharing mechanism - rather than employing independent Bi-LSTMs networks for Pinyin and Wubi, we let them share the same LSTMs with character embeddings. 

In Model-\uppercase\expandafter{\romannumeral3}, we feed character, Pinyin and Wubi embeddings sequentially into a stacked Bi-LSTMs network shared with the same parameters: 
\begin{equation}
\begin{aligned}
\begin{bmatrix}
\mathbf{h}_{3,c}^{(t)} \\[0.5em]
\mathbf{h}_{3,p}^{(t)} \\[0.5em]
\mathbf{h}_{3,w}^{(t)}
\end{bmatrix} &= \textrm{Bi}{\text -}\textrm{LSTMs} (\begin{bmatrix}
\mathbf{w}_{c}^{(t)} \\[0.5em]
\mathbf{w}_{p}^{(t)} \\[0.5em]
\mathbf{w}_{w}^{(t)}
\end{bmatrix}, \theta), \\
\mathbf{h}^{t} &= \mathbf{h}_{3,c}^{(t)} + \mathbf{h}_{3,p}^{(t)} + \mathbf{h}_{3,w}^{(t)},
\end{aligned}
\label{eq:model3}
\end{equation}
where $\theta$ denotes the shared parameters of Bi-LSTMs. Different from Eqn~\eqref{eq:model2}, there is only one shared Bi-LSTMs rather than three independent LSTM networks with more trainable parameters. In consequence, the shared Bi-LSTMs-CRF model can be trained more efficiently compared to Model-I and Model-II (extra FC-Layer expense).


Specifically, at each epoch, the parameters of three networks are updated based on unified sequential character, Pinyin and Wubi embeddings. The second LSTM network will share (or synchronize) the parameters with the first network before it begins the training procedure with Pinyin as inputs. In this way, the second network will take fewer efforts in refining the parameters
based on the former correlated embeddings. So does the third network (taking Wubi embedding as inputs). 

\section{Experimental Evaluations}
In this section, we provide empirical results to verify the effectiveness of multiple embeddings 
for CWS. Besides, our proposed Model-III can be trained efficiently (slightly costly than baseline) and obtain the state-of-the-art performance. 



\subsection{Experimental Setup}
To make the results comparable and convincing, we evaluate our models on SIGHAN 2005 \cite{DBLP:conf/acl-sighan/Emerson05} and Chinese Treebank 6.0 (CTB6) \cite{DBLP:journals/nle/XueXCP05} datasets, which are widely used in previous works. We leverage standard word2vec tool to train multiple embeddings. In experiments, we tuned the embedding size following \citet{DBLP:journals/corr/YaoH16} and assigned equal size (256) for three types of embedding. The 
number of Bi-LSTM layers is set as 3.

\subsection{Experimental Results}

\renewcommand{\arraystretch}{1.1}
\begin{table*}[htbp!]
	\begin{center}
		\setlength{\tabcolsep}{3pt}
		\begin{tabular}{cccccccccccccccccccc}
			\thickhline
			\multirow{2}{*}{\textbf{Models}} &  \multicolumn{3}{c}{\textbf{CTB6}} & &  \multicolumn{3}{c}{\textbf{PKU}} & &   \multicolumn{3}{c}{\textbf{MSR}} & & \multicolumn{3}{c}{\textbf{AS}} & &  \multicolumn{3}{c}{\textbf{CityU}} \\ \cline{2-20}
			& P & R & F && P & R & F && P & R & F && P & R & F && P & R & F \\ \hline
			baseline & 94.1 & 94.0 & 94.1 && 95.8 & 95.9 & 95.8 && 95.3 & 95.7 & 95.5 && 95.6 & 95.5 & 95.6 && 95.9 & 96.0 & 96.0 \\ 
			Model-I  & 94.9 & 95.0 & 94.9 && 95.7 & 95.7 & 95.7 && 96.8 & 96.6 & 96.7 && 96.6 & 96.5 & 96.5 && 96.7 & 96.5& 96.6 \\ 
			Model-II  & \textbf{95.4} & \textbf{95.3} & \textbf{95.4} && \textbf{96.3} & 95.7 & 96.0 && 96.6 & 96.5 & 96.6 && 96.8 & 96.5 & 96.7 && \textbf{97.2} & \textbf{97.0} & \textbf{97.1} \\
			Model-III & \textbf{95.4} & 95.0 & 95.2 && \textbf{96.3} & \textbf{96.1} & \textbf{96.2} && \textbf{97.0} & \textbf{96.9} & \textbf{97.0} && \textbf{96.9} & \textbf{96.8} & \textbf{96.9} && 97.1 & \textbf{97.0} & \textbf{97.1} \\ \thickhline
		\end{tabular}
	\end{center}
	\vspace{-2mm}
	\caption{Comparison of different architectures on five corpora. Bold font signifies the best performance in all given models. Our proposed multiple-embedding models result in a significant improvement compared to vanilla character-embedding baseline model.
	}
	\vspace{-2mm}
	\label{tab:architecture}
\end{table*}

\subsubsection*{Performance under Different Architectures}
We comprehensively conduct the analysis of three architecture proposed in Section~\ref{sec:architecture}. As illustrated in Table~\ref{tab:architecture}, considerable improvements are obtained by three multi-embedding models compared with our baseline model which only takes character embeddings as inputs. Overall, Model-III (shared Bi-LSTMs-CRF) achieves better performance even with fewer trainable parameters. 
\begin{table}[t]
	\begin{center}
		\setlength{\tabcolsep}{1.5pt}
		\begin{tabular}{c|cccc}
			\hline  Model &  PKU &  MSR &  AS &  CityU \\ \hline \hline
			\cite{DBLP:conf/acl/SunW12}  & 95.4 & 97.4 & - & - \\
			\cite{DBLP:conf/emnlp/ChenQZLH15} & 94.8 & 95.6 & - & - \\
			\cite{DBLP:conf/acl/ChenSQH17} & 94.3 & 96.0 & - & 94.8 \\
			\cite{ma2018state} & 96.1 & \textbf{97.4} & 96.2 & 97.2\\
            \hline \hline
			\cite{DBLP:conf/emnlp/ZhangWSM13}* & 96.1 & 97.4 & - & - \\
			\cite{DBLP:conf/emnlp/ChenQZLH15}* & 96.5 & 97.4 & - & - \\
			\cite{DBLP:conf/acl/CaiZZXWH17}* & 95.8 & 97.1 & 95.6 & 95.3\\
			\cite{2017arXiv1}* & 96.5 & 98.0 & - & -\\
			\cite{2017arXiv2}* & 96.0 & 97.9 & 96.1 & 96.9 \\ \hline \hline
			baseline & 95.8 & 95.5 & 95.6 & 96.0 \\
			ours (+PY)* & 96.0 & 96.8 & 96.7 & 97.0 \\
			ours (+WB) & \textbf{96.3} & 97.2 & 96.5 & \textbf{97.3} \\
			ours (+PY+WB)* & 96.2 & 97.0 & \textbf{96.9} & 97.1 \\
			\hline
		\end{tabular}
	\end{center}
	\vspace{-3mm}
	\caption{Comparison with previous state-of-the-art models on all four Bakeoff2005 datasets. The second block (*) represents allowing the use of external resources such as lexicon dictionary or trained embeddings on large-scale external corpora. Note that our approach \textbf{does not} leverage any external resources.} 
	\label{tab:compare4}
	\vspace{-5mm}
\end{table}

\subsubsection*{Competitive Performance}
To demonstrate the effectiveness of supplementary embeddings for CWS, we compare our models with previous state-of-the-art models.


Table~\ref{tab:compare4} shows the comprehensive comparison of performance on all Bakeoff2005 corpora. To the best of our knowledge, we have achieved the best performance on AS and CityU datasets (with F1 score 96.9 and 97.3 respectively) and competitive performance on PKU and MSR even if not leveraging external resources (e.g. pre-trained char/word embeddings, extra dictionaries, labeled or unlabeled corpora). It is worthy to notice that AS and CityU datasets are considered more difficult by researchers due to its larger capacity and higher out of vocabulary rate. It again verifies that Pinyin and Wubi embeddings are capable of decreasing mis-segmentation rate in large-scale data. 

\subsubsection*{Embedding Ablation}
We conduct embedding ablation experiments on CTB6 and CityU to explore the effectiveness of Pinyin and Wubi embeddings individually. 
As shown in Table~\ref{tab:ablation}, Pinyin and Wubi result in a considerable improvement on F1-score compared to vanilla single character-embedding model (baseline). 
Moreover, Wubi-aided model usually leads to a larger improvement than Pinyin-aided one. 

\renewcommand{\arraystretch}{1.1}
\begin{table}[htbp!]
    \centering
		\setlength{\tabcolsep}{3.1pt}
		\begin{tabular}{cccccccc}
			\thickhline
			\multirow{2}{*}{\textbf{Models}} &  \multicolumn{3}{c}{\textbf{CTB6}} & &  \multicolumn{3}{c}{\textbf{CityU}} \\ \cline{2-8}
			& P & R & F && P & R & F  \\ \hline
			baseline & 94.1 & 94.0 & 94.1 && 95.9 & 96.0 & 96.0 \\ 
			IO + PY & 94.6 & 94.9 & 94.8 && 96.8 & 96.4 & 96.6  \\
			IO + WB & 95.3 & \textbf{95.4} & 95.3 && \textbf{97.3} & \textbf{97.3} & \textbf{97.3} \\
			Model-II  & \textbf{95.4} & 95.3 & \textbf{95.4} && 97.2 & 97.0 & 97.1 \\ \thickhline
		\end{tabular}
	\vspace{-2mm}
	\caption{Feature ablation on CTB6 and CityU. IO + PY and IO + WB denote injecting Pinyin and Wubi embeddings separately under Model-II.}
	\vspace{-1mm}
	\label{tab:ablation}
\end{table}



\subsubsection*{Convergence Speed}

\begin{table}[htb]
    \begin{center}
		\setlength{\tabcolsep}{5pt}
		\begin{tabular}{c|cc}
			\hline  Model & Time (batch) & Time (P-95\%)  \\ \hline
			baseline & 1 $\times$ & 1 $\times$\\
			Model-I & 2.61 $\times$ & 2.51 $\times$\\
			Model-II & 1.03 $\times$ & 1.50 $\times$\\
			Model-III & \textbf{1.07} $\times$ & \textbf{1.04} $\times$\\ 
			\hline
		\end{tabular}
	\end{center}
	\vspace{-3mm}
	\caption{Relative training time on MSR. (a) averaged training time per batch; (b) convergence time, where above 95\% precision is considered as convergence.}
	\vspace{-5mm}
	\label{tab:comparespeed}
\end{table}

To further study the additional expense after incorporating Pinyin and Wubi, we record the training time (batch time and convergence time in Table~\ref{tab:comparespeed}) of proposed models on MSR. Compared to the baseline model, it almost takes the same training time ($1.07\times$) per batch and convergence time ($1.04\times$) for Model-III. By contrast, Model-II leads to slower convergence ($1.50\times$) in spite of its lower batch-training cost.
In consequence, we recommend Model-III in practice for its high efficiency. 

\section{Related Work}
Since \citet{xue2003chinese}, researchers have mostly treated CWS as a sequence labeling problem. Following this idea, great achievements have been reached in the past few years with the effective embeddings introduced and powerful neural networks armed.

In recent years, there are plentiful works exploiting different neural network architectures in CWS. Among these architectures, there are several models most similar to our model: Bi-LSTM-CRF \cite{DBLP:journals/corr/HuangXY15}, Bi-LSTM-CRF \cite{lample2016neural,101007}, and Bi-LSTM-CNNs-CRF \cite{DBLP:conf/acl/MaH16}.

\citet{DBLP:journals/corr/HuangXY15} was the first to adopt Bi-LSTM network for character representations and CRF for label decoding. 
\citet{lample2016neural} and \citet{101007} exploited the Bi-LSTM-CRF model for named entity recognition in western languages and Chinese, respectively.
Moreover, \citet{101007} introduced radical-level information that can be regarded as a special case of Wubi code 
in our model. 

\citet{DBLP:conf/acl/MaH16} proposed to combine Bi-LSTM, CNN and CRF, which results in faster convergence speed and better performance on POS and NER tasks. In addition, their model leverages both the character-level and word-level information. 

Our work distinguishes itself by utilizing multiple dimensions of features in Chinese characters. With phonetic and semantic meanings taken into consideration, three proposed models achieve better performance on CWS and can be also adapted to POS and NER tasks. In particular, compared to radical-level information in \cite{101007}, Wubi Input encodes richer structure details and potentially semantic relationships. 

Recently, researchers propose to treat CWS as a word-based sequence labeling problem, which also achieves competitive performance \cite{DBLP:conf/acl/ZhangZF16,DBLP:conf/acl/CaiZ16,DBLP:conf/acl/CaiZZXWH17,DBLP:journals/corr/YangZD17}. Other works try to introduce very deep networks \cite{2017arXiv1} or treat CWS as a gap-filling problem \cite{2017arXiv2}. We believe that proposed linguistic features can also be transferred into word-level sequence labeling and correct the error. In a nutshell, multiple embeddings are generic and easily accessible, which can be applied and studied further in these works.

\section{Conclusion}
\label{sec:conclusion}
In this paper, we firstly propose to leverage phonetic, structured and semantic features of Chinese characters by introducing multiple character embeddings (\textit{Pinyin} and \textit{Wubi}). We conduct a comprehensive analysis on why Pinyin and Wubi embeddings are so essential in CWS task and could be translated to other NLP tasks such as POS and NER. Besides, we design three generic models to fuse the multi-embedding and produce the start-of-the-art performance in five public corpora. In particular, the shared Bi-LSTM-CRF models (Model III in Figure~\ref{fig:architecture}) could be trained efficiently and produce the best performance on AS and CityU corpora. In future, the effective ways of leveraging hierarchical linguistic features to other languages, NLP tasks (e.g., POS and NER) and refining mis-labeled sentences merit further study. 




\section*{Acknowledgement}
This research work has been partially funded by the National Natural Science Foundation of China (Grant No.61772337, U1736207), and the National Key Research and Development Program of China NO.2016QY03D0604.

\bibliography{acl2019}
\bibliographystyle{acl_natbib}

\appendix

\clearpage

\end{CJK*}
\end{document}